\documentclass[letterpaper, 10 pt, conference]{ieeeconf}  % Comment this line out if you need a4paper

\IEEEoverridecommandlockouts                              % This command is only needed if 
\usepackage{graphicx} % for pdf, bitmapped graphics files
\usepackage{amsmath} % assumes amsmath package installed
\usepackage{amssymb}  % assumes amsmath package installed

\usepackage{stfloats}
\usepackage[font=footnotesize]{caption}
\usepackage{balance}
\usepackage{nameref}
\usepackage{siunitx}
\usepackage{booktabs}
\usepackage{hyperref}
\usepackage{pgfplots}
\usepackage{tikzscale}
\pgfplotsset{compat=newest}
\newlength\figureheight
\newlength\figurewidth 

\usepackage[noadjust]{cite}

\usepackage{color}
\usepackage{import}
\newcommand{\executeiffilenewer}[3]{%
	\ifnum\pdfstrcmp%
		{\pdffilemoddate{#1}}%
		{\pdffilemoddate{#2}}%
		>0%
		{\immediate\write18{#3}}%
	\fi%
}

\title{\LARGE \bf
	Object Detection on Dynamic Occupancy Grid Maps Using \\
	Deep Learning and Automatic Label Generation
}

\author{Stefan Hoermann, Philipp Henzler, Martin Bach, and Klaus Dietmayer
\\
Institute of Measurement, Control, and Microtechnology, Ulm University, Germany	
}

\def\mysection#1#2{\section{#1}\label{sec:#2}}
\def\mysubsection#1#2{\subsection{#1}\label{sec:#2}}

\usepackage{soul}
\usepackage{amsmath}
\usepackage{amssymb}
\usepackage{microtype}
\usepackage{wrapfig}
\usepackage{todonotes}
\usepackage{graphicx}

\newcommand{\eg}{e.\,g.\ }
\newcommand{\ie}{i.\,e.\ }
\newcommand{\etal}{et~al.\ }

\newcommand{\refFig}[1]{Fig.~\ref{fig:#1}}

\soulregister\ref7
\soulregister\cite7
\soulregister\refFig7
\soulregister\cite7
\soulregister\ref7
\soulregister\pageref7
\soulregister\shortcite7
\soulregister\eg0
\soulregister\ie0
\soulregister\etal0

\DeclareGraphicsExtensions{.png,.jpg,.pdf,.ai,.psd}
\begin{document}

\bstctlcite{IEEEexample:BSTcontrol} 

\def\cred{\textcolor{red}}
\def\cblue{\textcolor{blue}}
\def\cgreen{\textcolor{green}}

% Abkürzungen
\def\etal{et\:al.\ }
\def\DOGMA{DOGMa}

% % Symbols definition
% grid cell
\def\GMcell{c}
\def\GMtimestep{t}
\def\GMnuminliers{n}

% Channel
%\def\OUTchannel{\mathcal{T}}
\def\GMchannels{\Omega}
\def\GMwidth{W}
\def\GMheight{H}

\def\Anchor{\alpha}
\def\ANCnumshapes{C_{s}}
\def\ANCnumorientations{C_{\ANCori}}
\def\ANCnumanchors{C_{\Anchor}}
\def\ANCnumlengths{C_{\ANClength}}
\def\ANCnumwidths{C_{\ANCwidth}}
\def\ANCTolerance{\delta}

\def\OBJwidth{w}
\def\OBJlength{l}
\def\OBJori{\phi}
\def\OBJaspect{a}

\def\ANCwidth{\OBJwidth}
\def\ANClength{\OBJlength}
\def\ANCori{\OBJori}
\def\ANCapsect{\OBJaspect}
\def\ANCscale{\ANClength}

\def\ANCdWrel{\Delta{\ANCwidth}}
\def\ANCdLrel{\Delta{\ANClength}}
\def\ANCdOriRel{\Delta{\ANCori}}

\def\Occupied{\mathrm{O}}
\def\MassFree{M_\mathrm{F}}
\def\MassOcc{M_\mathrm{O}}
\def\East{\mathrm{E}}
\def\North{\mathrm{N}}

\def\static{\mathrm{s}}
\def\dynamic{\mathrm{d}}

\def\IoUthreshold{\gamma}

\def\Netout{\hat{y}}
\def\Label{y}

\def\NetoutIoU{\Netout^{(\mathrm{IoU})}}
\def\NetoutDW{\Netout^{(\ANCdWrel)}}
\def\NetoutDL{\Netout^{(\ANCdLrel)}}
\def\NetoutDOri{\Netout^{(\ANCdOriRel)}}

\def\LabelIoU{\Label^{(\mathrm{IoU})}}
\def\LabelDW{\Label^{(\ANCdWrel)}}
\def\LabelDL{\Label^{(\ANCdLrel)}}
\def\LabelDOri{\Label^{(\ANCdOriRel)}}

\def\NetoutIoUMax{\hat{\mathbf{A}}}
\def\LabelIoUMax{\mathbf{A}}

\def\FocusingParam{f}
\def\GainParam{\lambda_\mathrm{I}}

\twocolumn[{%
\renewcommand\twocolumn[1][]{#1}%
	\maketitle
	\begin{center}
		\centering
		\includegraphics[width=\textwidth]{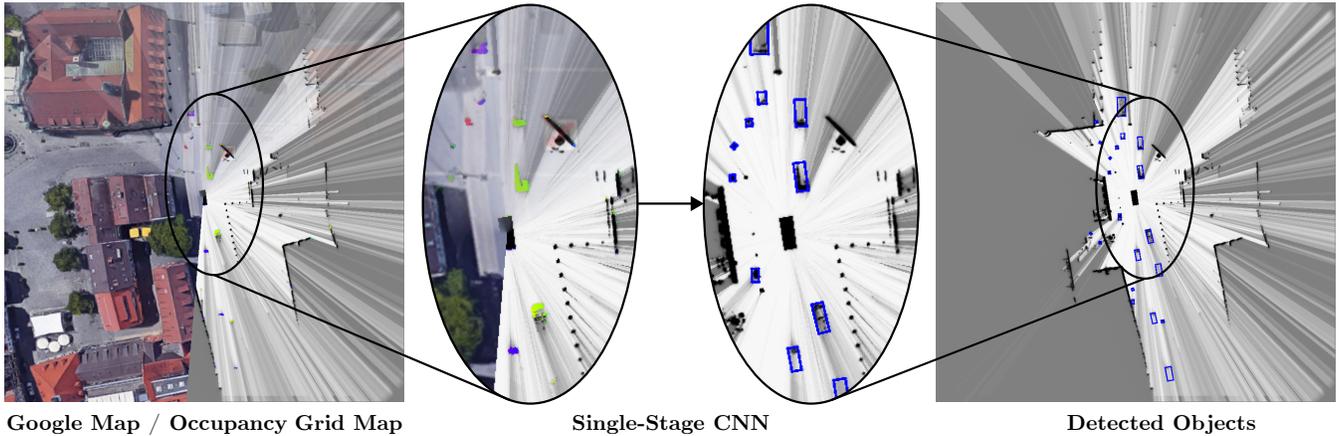}
		\captionof{figure}{Object detection using deep learning and grid fusion.}
		\label{fig:teaser}
	\end{center}%
}]

%\saythanks

\thispagestyle{empty}
\pagestyle{empty}

%%%%%%%%%%%%%%%%%%%%%%%%%%%%%%%%%%%%%%%%%%%%%%%%%%%%%%%%%%%%%%%%%%%%%%%%%%%%%%%%
\begin{abstract}

We tackle the problem of object detection and pose estimation in a shared space downtown environment.
For perception multiple laser scanners with 360$^\circ$ coverage were fused in a dynamic occupancy grid map (\DOGMA).
A single-stage deep convolutional neural network is trained to provide object hypotheses comprising of shape, position, orientation and an existence score from a single input \DOGMA.
Furthermore, an algorithm for offline object extraction was developed to automatically label several hours of training data.
The algorithm is based on a two-pass trajectory extraction, forward and backward in time.
Typical for engineered algorithms, the automatic label generation suffers from misdetections, which makes hard negative mining impractical.
Therefore, we propose a loss function counteracting the high imbalance between mostly static background and extremely rare dynamic grid cells.
Experiments indicate, that the trained network has good generalization capabilities since it detects objects occasionally lost by the label algorithm.
Evaluation reaches an average precision (AP) of $\SI{75.9}{\percent}$.

\end{abstract}

%%%%%%%%%%%%%%%%%%%%%%%%%%%%%%%%%%%%%%%%%%%%%%%%%%%%%%%%%%%%%%%%%%%%%%%%%%%%%%%%

% *** Section 1 - Introduction ***
\mysection{Introduction}{sec:introduction}

On the path to autonomous driving, a thoroughly modeled environment is essential for high-level software modules like behavior and trajectory planning \cite{kunz2015}.
Two environment representation strategies are commonly used: object-model-based and object-model-free.
We refer to an object model in terms of a vector containing the size and pose, while dynamics, existence probability and the according covariance matrix can also be included.
\emph{Object-model-free} grid maps gained great success fusing raw sensor data in one environment representation.
The task of object hypotheses generation is intentionally avoided \cite{Elfes1989, NUSS2015, NegreDynGM2014}.
Instead, grid fusion aims at estimating the occupancy probability and dynamic states at independent, discretized locations in the environment.
Thus, the strength of multiple sensors can be fused in a single dynamic occupancy grid map (\DOGMA) \cite{NUSS2016}, without the need to decide what object type caused the measurement.
However, modeled objects are fundamentally required for many applications such as decision-making \cite{ulbrich2013decmak, Brechtel2014decmak}.
\emph{Object-model-based} tracking aims for extracting moving objects represented as a list or distribution of object vectors \cite{Reuter15GGIWLMB,Danzer16ALMB}, while stationary objects are widely ignored.
Initializing objects and associating measurements is one of the most critical tasks in object tracking.
While sophisticated shape models were designed to associate measurements to objects \cite{TingYuan2017GMLaserShape, Bharanidhar2016LShapes, Roos2016RadarVehicleOriEstimation}, background models are mostly rudimentary, e.g. assuming uniform distributed clutter measurement.
Consequentially, objects are falsely detected as positive.
Convolutional neural networks (CNNs), alternatively, are known for their capability to exploit context information, or in other words, establishing an intrinsic background model.

In this work we present a learning based approach to finding objects in terms of width, length, orientation, and position in a \DOGMA\ as illustrated in Fig.~\ref{fig:teaser}.
The left side depicts a top view satellite image from Google Maps with a fading overlay of a \DOGMA. 
The right side depicts detected objects as blue rectangles.
%

%%%%%%%%%%%%%%%%%%%%%%%%%%%%%%%%%%%%%%%%
%% Labelling %%
%%%%%%%%%%%%%%%%%%%%%%%%%%%%%%%%%%%%%%%%

Whereas the extensive task of manual labeling in learning applications is a main drawback, we propose a fully automated approach to extract labels of moving objects.
By collecting object data over time and feeding gained information back to earlier time steps where important information like the object shape wasn't observed yet, an acausal object extraction algorithm is presented for offline label generation.
%
%%%%%%%%%%%%%%%%%%%
%% Loss Function %%
%%%%%%%%%%%%%%%%%%%

The remaining paper is organized as follows:
Related work is reviewed in Section~\ref{sec:related_work}.
Details about dynamic occupancy grid maps, used as single input to a deep convolutional neural network (CNN), are given in Section~\ref{sec:dogma}.
An object extraction algorithm used to generate hours of training data without manual labeling is introduced in Section~\ref{sec:label_generation}.
Section~\ref{sec:network_output} explains the network architecture and its output which is based on the concept of 'anchors', where the network predicts the best fit within a set of default bounding boxes plus the offset to the true object.
Section~\ref{sec:loss_function} proposes a loss function, particularly adapted to the extremely imbalanced character of the data.
%
%The designed loss function is introduced in Section~\ref{sec:loss_function}.
%
%\cred{It is important for successful training, due to the extreme imbalanced character of our data, which is thoroughly examined in} Section~\ref{sec:dataset_training}.
Subsequently, the data itself is examined in Section~\ref{sec:dataset_training}.
Experiments, showing the precision recall behavior of the detection network as well as the bounding box error objects, are carried out in Section~\ref{sec:results}, followed by conclusions in Section~\ref{sec:conclusions}.

% *** Section 2 - related work ***
\mysection{Related Work}{related_work}
A common object tracking approach unites raw measurements by box fitting and tracking these boxes considered as single pseudo measurements.
While sensor fusion and tracking approaches are advanced and theoretically supported, object detection or initialization is reasonably engineered.
Hand engineered object detection based on box fitting with L-shapes in laser \cite{Munz2009Lshapes,Bharanidhar2016LShapes} or radar measurements \cite{Roos2016RadarVehicleOriEstimation}, suffers from limiting assumptions and simplifications regarding sensor, object, and environment features.
Commonly, heuristic parameter tuning is required, e.g. to adjust the measurement noise and clutter assumptions.
A data driven alternative to box fitting is proposed by Scheel and Dietmayer \cite{Scheel2017learnedRadarModel}, where the radar measurement model of a car is learned and can be probabilistically conditioned on the perspective.
So far, however, the aforementioned approach only focuses on cars.
%

%%%%%%%%%%%%%%%%%%%%%%%%%%%%
%% DOGMA Object Detection %%
%%%%%%%%%%%%%%%%%%%%%%%%%%%%
%
Detecting objects in grid maps widely decouples sensor fusion from object tracking, i.e. performing object detection after dynamic grid mapping.
To generate spatially extended object models, highly engineered methods were proposed to find cell clusters representing an object \cite{SchuetzDietmayer2014GMObjects, TanzmeisterGridTracking2017, SteyerTanzmeister2017GMObjectTracking}.
Experiments showing extracted and tracked objects seem promising, however, the engineered initialization requires easy separable cells with small velocity variance. 
Generally, object detection in grid maps suffers from corrupted object silhouettes, occlusions, and false velocity estimates in static regions.
We claim, that a CNN can deal with these corner cases.

%%%%%%%%%%%%%%%%%%%%%%%%%%%%
%% Learning on Grid Maps? %%
%%%%%%%%%%%%%%%%%%%%%%%%%%%%

A \DOGMA\ provides a neural network friendly representation of the entire environment and its dynamics. 
Piewak \etal \cite{Piewak2017GMObjects} trained a neural network to reduce false positive velocity estimation in a \DOGMA\ sequence. 
In particular, their approach refers to a pixel-wise classification task to determine whether a cell in a \DOGMA\ is dynamic or static. 
Yet, clustering is still necessary to obtain objects.
In contrast, our approach directly predicts objects with position, shape and orientation.
In our previous work \cite{hoermann2017GMPred}, a network was trained to separate static regions in a \DOGMA\ and predict future cell occupancy caused by dynamic regions.
Similarly, Dequaire~\etal\cite{dequaire2016deep} propose an end-to-end recurrent neural network for object tracking. 
Again, their approach infers pixel states while our results, in this work, are bounding boxes.
%

%%%%%%%%%%%%%%%%%%%%%%%%%%%%%%%%%%%%%%%%
%% Learned Object Detection Allgemein %%
%%%%%%%%%%%%%%%%%%%%%%%%%%%%%%%%%%%%%%%%

%
Object detection in general, or more specifically in images, is a very active research area.
Recent approaches like Fast R-CNN \cite{FastRCNN}, Faster R-CNN \cite{FasterRCNN}, or Mask R-CNN \cite{MaskRCNN}, are based on two stages. 
The first stage delivers a region of interest (ROI) in an image, e.g. provided by a Region Proposal Network (RPN) \cite{FasterRCNN}.
The second stage is applied on the ROIs to predict a classification and ROI offset.
The recent Mask R-CNN \cite{MaskRCNN} implements a similar two-stage procedure but adds a binary mask output for every ROI for additional object segmentation.

Other approaches, like YOLO \cite{YOLO} and SSD \cite{SSD}, employ a deep convolutional neural network (CNN) only in a single stage. 
The former tries to infer object bounding boxes by regression, which however lacks of position accuracy and couples bounding box regression with classification.
On the other hand, SSD follows the strategy of anchors, i.e. default boxes defined by aspect ratio and scale, to find bounding boxes of classified objects.
In addition to classifying default boxes, the shape offset is predicted by the neural network for each default box.
% ZITAT: " For each default box, we predict both the shape offsets and the confidences for all object categories "
%
To counteract the imbalance in the data with respect to default boxes (most boxes have negative decision), only the top 3 negative decisions are sampled during training.
We also follow the approach of default boxes but additionally add object orientation as an attribute of an anchor.
An optimization approach is used on a large dataset, to choose the set of anchors.

Focal Loss \cite{FocalLoss} investigated the great success of two-stage approaches and found, that the extreme imbalance between background and object pixels was compensated by the ROI extraction.
Using a novel loss function, which considers this imbalance implicitly, enabled a one-stage network to gain even better performance. 
% ZITAT: RetinaNet is able to match the speed of previous one-stage detectors while surpassing the accuracy of all existing state-of-the-art two-stage detectors.
%
We also faced this problem in our previous work \cite{hoermann2017GMPred} when dealing with the high imbalance between static and dynamic cells in a \DOGMA and proposed a pixel balancing loss function which we adopt in this work.
%

% *** Section - Dynamic Occupancy Grid Map ***
\mysection{Filtered Dynamic Input}{dogma}
\begin{figure}[t]
	\centering
	\vspace{2mm}
	\def\shroistartX{315}
	\def\shroistartY{420}
	\def\shroiWidth{187}
	\def\shroiHeight{105}
	\def\shifigurewidth{0.18}
	\includegraphics[width=0.9\linewidth]{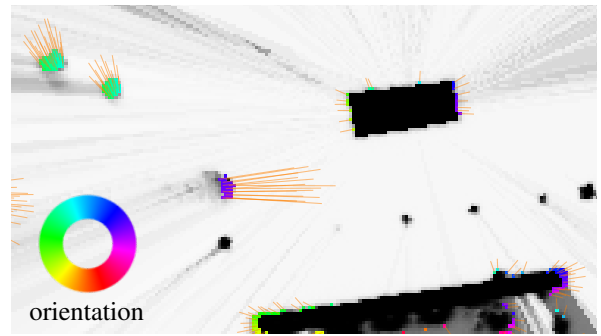}
	\caption{Excerpt of a \DOGMA. 
		Orange lines indicate the estimated cell velocity, while the cell color indicates the movement orientation.
		Grayscale indicates the cell occupancy probability $P_\Occupied$.}
	\label{fig:gridmapexample}
\end{figure}
We use the \DOGMA\ from Nuss \etal\cite{NUSS2016}.
Fig.~\ref{fig:gridmapexample} shows an excerpt of a \DOGMA\ created from multiple laser scanners.
The perceived environment is spatially discretized in grid cells $\GMcell$ at positions $(\East, \North)$, indicating east and north, respectively.
A particle-filter-based velocity estimation augments the classical static occupancy grid with dynamic states.
Occupancy estimation is based on Dempster Shafer \cite{dempster2008generalization}.
A cell contains the channels $\GMchannels = \left\{ \MassOcc, \MassFree, v_\East, v_\North, \sigma^2_{v_\East}, \sigma^2_{v_\North}, \sigma^2_{v_\East, v_\North} \right\}$ with Dempster-Shafer masses for free space $\MassFree \in \left[0,1\right]$ and occupancy $\MassOcc \in \left[0,1\right]$, the velocity pointing east $v_\East$ and north $v_\North$, as well as the velocity variances and covariance.
The occupancy probability is calculated by $P_\Occupied = 0.5\cdot\MassOcc + 0.5\cdot(1-\MassFree)$ where a high $P_\Occupied$ refers to a dark pixel in Fig. \ref{fig:gridmapexample}.
$P_\Occupied(\East, \North, \GMtimestep) := P_\Occupied(\GMcell, \GMtimestep)$ denotes the occupancy probability at a grid cell $\GMcell$ and sequence time step $\GMtimestep$.
The \DOGMA\ data is provided in $\mathbb{R}^{\GMwidth \times \GMheight \times |\GMchannels|}$ with the spatial width $\GMwidth$ and height $\GMheight$ pointing east and north, respectively.

A key assumption for efficient processing is the independence of single cells.
In consequence, as observable in Fig. \ref{fig:gridmapexample}, borders of walls and static cars tend to have false velocity estimates.
This causes simple clustering to fail, while a convolutional neural network can be trained to consider context.
%

% *** Section - ***
%\input{doc/sec_generating_labels}
%\mysection{Automatic Label Generation}{label_generation}
\begin{figure*}[t]
	\vspace{2mm}
	\includegraphics[width=\linewidth]{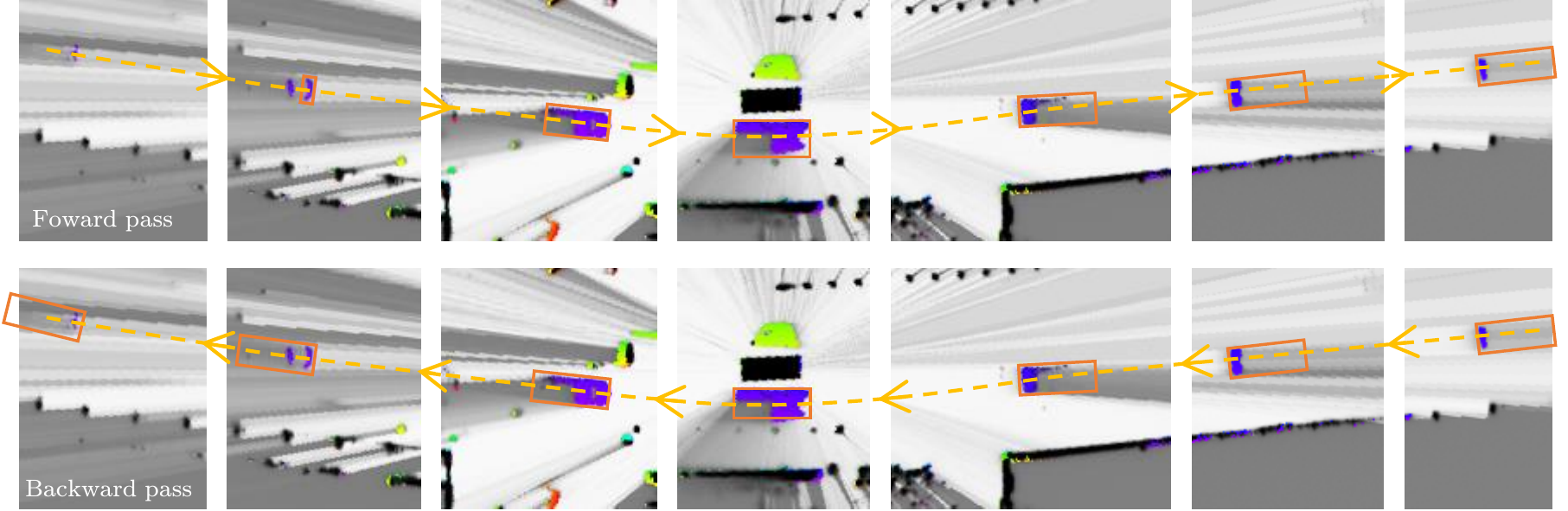}
	\caption{Schematic of the two-pass label extraction algorithm: In the forward pass, well detectable objects are tracked through the sequence. In the backward path, the object shape and pose is refined and even a corrupted object silhouette can be used to fit the object bounding box.}
	\label{fig:trajectoryexport}
\end{figure*}

\section{Automatic Label Generation}
\label{sec:label_generation}
Highly engineered object detection in grid maps, i.e. estimation of position, width, length and orientation, is examined in literature \cite{SteyerTanzmeister2017GMObjectTracking,SchuetzDietmayer2014GMObjects}.
The algorithms are relatively restricted due to application constraints, e.g. real time processing.
Our application, in contrast, is offline label extraction which allows advantageous reduction of restrictions.
For example, the algorithm ignores causality: it runs forward and backward in time, and makes use of preprocessing, e.g., spatial and temporal data smoothing using multidimensional Gaussian kernels.
Furthermore, post processing is used to refine trajectories and identify outliers on a pixel and trajectory level. 
Like most engineered algorithms, a relatively high false negative rate can be assumed, compared to human object detection.
However, considering these circumstances during network training, utilizing a learning approach could lead to better generalization.
Hence, endless labeled sequences are imaginable.

%%%%%%%%%%%%%%%%%%%%%%%%%%%%%%
%% Object tracking key idea %%
%%%%%%%%%%%%%%%%%%%%%%%%%%%%%%
%
Fig. \ref{fig:trajectoryexport} illustrates the extraction of a vehicle.
The extracted bounding box is drawn in orange at different time steps, while the cell cluster appears purple.
The vehicle approaches and passes the ego-vehicle, which appears as a black, filled rectangle in the center.
The purple object silhouette is corrupted due to (self) occlusion and particle filter convergence.
When objects enter the field of view, they are not visible very clearly in a \DOGMA, and their silhouette grows when they get closer to the ego vehicle.
When the object passed the ego-vehicle, the visible object silhouette shrinks.
Successively, the front, side and back of the vehicle are visible and exhibit a rectangular object shape. 
In the forward pass (top row in Fig. \ref{fig:trajectoryexport}) object tracking is initialized when it is very certain that a cell belongs to a moving object, i.e. high $P_\Occupied(\GMcell, \GMtimestep)$, low velocity variance, and high velocity magnitude.
After the vehicle leaves the field of view, the backward pass is initialized (bottom row), refining the object pose and extent, and detecting objects in time steps before the object was initialized.
At each time step in the sequence, an object corner point expected to be visible, named reference point, is found considering object orientation, size, position, as well as occlusions in the line of sight to the rectangle corner points.
Thus, a bounding box can be constructed starting from the reference point even at corrupted or partially occluded object silhouettes in a far sensing region.

The silhouette clustering is straight forward, based on connected components, i.e. connected cells with similar $P_\Occupied$ and velocity.
However, some extensions are described in the following.
%
%%%%%%%%%%%%%%%%%%%
%% Preprocessing %%
%%%%%%%%%%%%%%%%%%%
%
To limit cell clusters, boundary cells are calculated ideally limiting the object silhouette at a rise or slope of the smoothed occupancy probability $P_\Occupied(\East, \North, \GMtimestep)$.
For this, the first and second spatial derivative of $P_\Occupied(\East, \North, \GMtimestep)$ is calculated to find inflection points.
This is in particular useful when objects are close to other objects or static regions.
The found object silhouette is predicted to the next \DOGMA\ time step using velocity statistics from the spatial velocity distribution as well as the velocity covariances of single cells.
Cells covered by the predicted silhouette and fitting best to the velocity profile are chosen to start a new connected component search. 
The number of start cells is scaled by the expected object silhouette size to include about $1$ cell per \SI{0.5}{m^2}.
It is assumed, that the new connected component contains outliers.
Therefore, velocity statistics of $\GMnuminliers$ inlier cells with the least velocity variance and highest $P_\Occupied$ are chosen to establish new object cell statistics.
Remaining cells of the connected component are considered as outliers if they are outside a $2\sigma$ bound.
$\GMnuminliers$ is the number of cells included in the previous extracted silhouette if the silhouette grows, or half of new initial cells otherwise.
%

%%%%%%%%%%%%%%%%%%%%%
%% Post Processing %%
%%%%%%%%%%%%%%%%%%%%%
%
In post processing, the extracted trajectories are smoothed using spline fitting.
Trajectories with unreasonable motion are rejected.
Furthermore, open street map \cite{OpenStreetMap} is used to eliminate static areas falsely detected as objects and mirrored objects in glass fronts of buildings.

A main drawback of the algorithm is, that if an object is lost in a late stage of the trajectory, it is hard to resume tracking. 
The same applies in an early stage in the backward pass.
Therefore, the labeled data tends to contain more missing labels than false positives.
Since we are aware of this problem, it can be considered when training the network.

% *** Section - ***
\mysection{Network Output and Architecture}{network_output}

\begin{figure}[b]
	\centering
	\includegraphics[width=\linewidth]{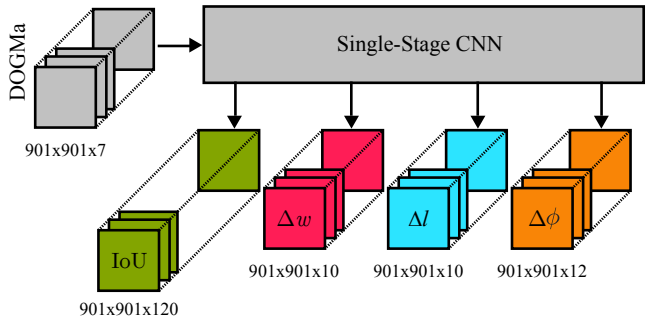}
	\caption{Network input and output. The network predicts scores (IoU) of default 'anchor' boxes and feature offsets to the object bounding box.}
	\label{fig:netout}
\end{figure}

We chose a simple encoder-decoder network structure with skip connections, inspired by \cite{noh2015deconvnet}. 
We employ a pixel-to-pixel structure yielding equal input and output resolution.
Instead of fully regressing bounding boxes, we follow the strategy to classify a limited number of rotated default boxes (anchors) and additionally regress their offset to the ground truth box. 
Anchors are defined by triples $\left(\ANCwidth, \ANClength, \ANCori \right)$, denoting the width, length and orientation, respectively.
A more intuitive representation is $\left(\ANCapsect, \ANClength, \ANCori \right)$, where $\ANCapsect = \frac{\ANCwidth}{\ANClength}$ refers to the aspect and $\ANClength$ can be interpreted as a scale.
We refer to $\left(\ANCapsect, \ANClength \right)$ as the shape.
In the following, we distinguish between label and network prediction by using a $\wedge$ on top of a label symbol.

The network output is illustrated in Fig.~\ref{fig:netout}.
It is trained to produce four different outputs: The anchor score $\NetoutIoU$, the width offset $\NetoutDW$, the length offset $\NetoutDL$ and the orientation offset $\NetoutDOri$.
$\NetoutIoU$ is encoded as the intersection over union between default box and true box, commonly used to compare similarity of bounding boxes \cite{Everingham2010}.
$\NetoutDW$ and $\NetoutDL$ are provided relative to the anchor width and length, respectively, in order to gain similar values for all anchors.
The orientation offset $\NetoutDOri$ is scaled to $\pi$.

With $\ANCnumshapes$ default shapes and $\ANCnumorientations$ default orientations,  $\ANCnumanchors = \ANCnumshapes \cdot \ANCnumorientations$ anchors $\Anchor$ are defined.
From this follows that $\NetoutIoU \in \mathbb{R}^{\GMwidth\times\GMheight\times\ANCnumanchors}$ couples the three features $\ANClength$, $\ANCwidth$ and $\ANCori$ in a single prediction for each grid cell $\GMcell \in \left\{1, ..., \GMwidth\cdot\GMheight\right\}$.
Thus, for every grid cell the score of each default bounding box is provided, assuming the cell is the center of the box.
It is important to note that we chose not to train for a binary decision, but regressing the IoU between anchor and ground truth box.
That way, the box fitting is essentially discretized to decide for a default box, while the decision itself is made via regression of the IoU.

An alternative to coupling box features in one prediction is to use independent box feature outputs for $\ANCwidth$, $\ANClength$, and $\ANCori$.
Although this could reduce the output dimension to the sum $\ANCnumwidths + \ANCnumlengths + \ANCnumorientations$, it also allows for unreasonable box results, e.g. estimating the length of a truck, the width of a bike but the orientation of a pedestrian.
However, shape offset and orientation offset are assumed to be independent. 
In fact, the orientation offset is constant for default shapes with equal orientation.
Consequently, the shape offset outputs are provided with $\ANCnumshapes$ and the orientation output with $\ANCnumorientations$ channels, which leads to $\NetoutDW, \NetoutDL \in \mathbb{R}^{\GMwidth\times\GMheight\times\ANCnumshapes}$ and $\NetoutDOri \in \mathbb{R}^{\GMwidth\times\GMheight\times\ANCnumorientations}$.
Offset regression is trained for all default boxes, no matter if the according anchor box fits best or not at all to the ground truth object.

\mysubsection{Anchor Selection}{choosing_anchors}

\begin{figure}[t]
\vspace{1.5mm}
\centering
\includegraphics[clip, width = 8.4cm]{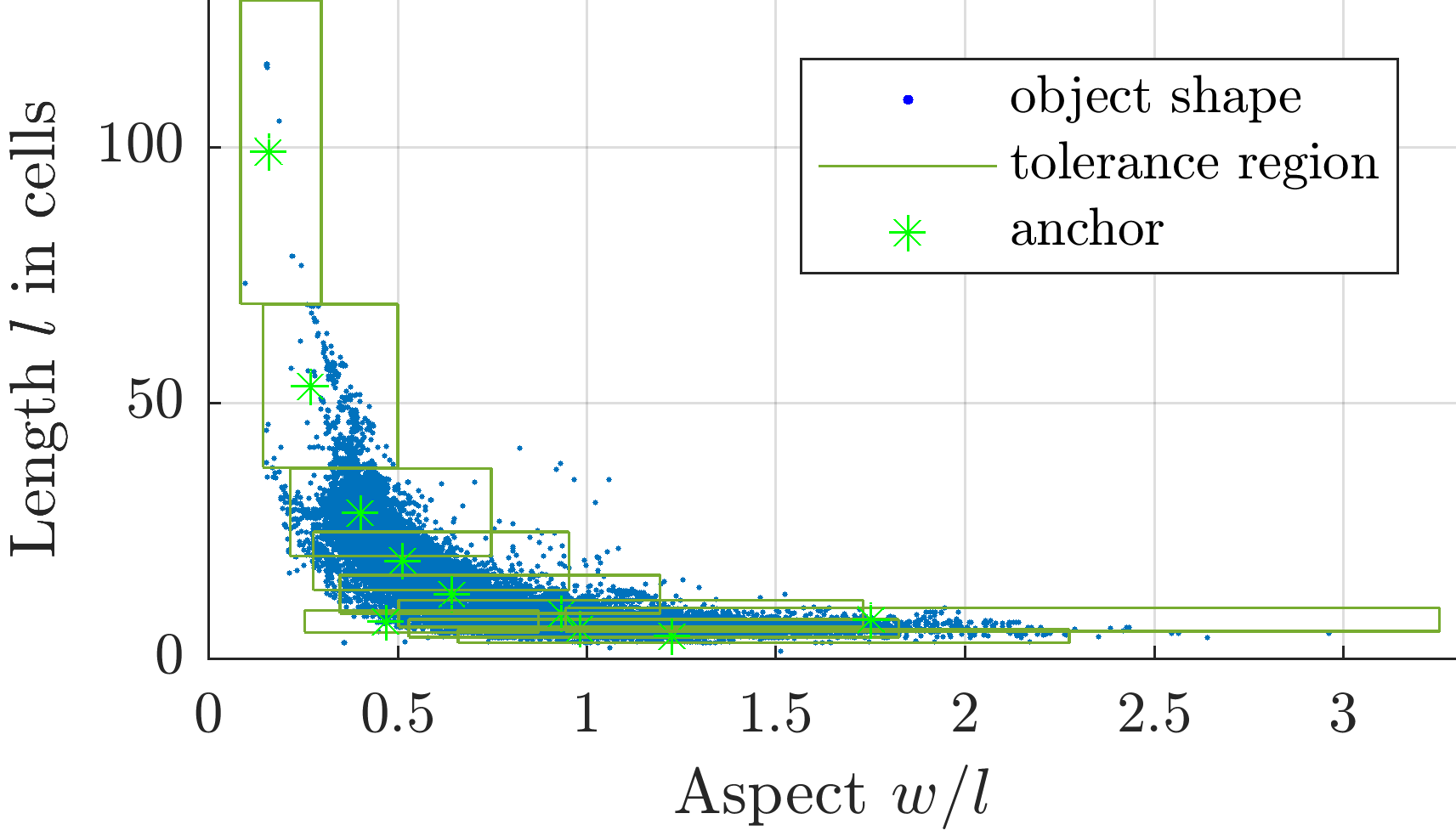}
\caption{Chosen anchors (green stars) over ground truth box shapes (blue dots). The rectangles illustrate the tolerance region (green rectangles).}
\label{fig:anchors}
\end{figure}

We acquired numerous rotated rectangle labels with width $\OBJwidth$, length $\OBJlength$ and orientation $\OBJori$.
While the orientation is considered uniformly distributed, the anchor shapes should only cover a sparse set of reasonable aspects and scales.
Therefore, we define anchor orientation and shape independently where $\ANCnumorientations=12$ anchor orientations were defined equally distributed in  $[0,2\pi)$ and $\ANCnumshapes=10$ default rectangle shapes were found by optimization.
Anchor shape optimization aims to cover most label shapes within a preliminarily defined offset tolerance of $\ANCTolerance = \SI{30}{\percent}$.
Since the algorithm operates in $(\ANCapsect, \ANClength)$ space, we define tolerance ranges by $\ANClength_\mathrm{min} = \ANClength\cdot(1 - \ANCTolerance)$, $\ANClength_\mathrm{max} = \ANClength\cdot(1 + \ANCTolerance)$, 
%
%$\ANCapsect_\mathrm{min} = \frac{\ANCapsect\cdot\ANClength\cdot(1-\ANCTolerance)}{\ANClength_\mathrm{max}}$ 
$\ANCapsect_\mathrm{min} = \ANCapsect\frac{\ANClength_\mathrm{min}}{\ANClength_\mathrm{max}}$ 
and
%
%$\ANCapsect_\mathrm{max} = \frac{\ANCapsect\cdot\ANClength\cdot(1+\ANCTolerance)}{\ANClength_\mathrm{min}}$.
$\ANCapsect_\mathrm{max} =\ANCapsect \frac{\ANClength_\mathrm{max}}{\ANClength_\mathrm{min}}$.
The algorithm operates in the following manner:
A 2D histogram over $\OBJaspect$ and $\OBJlength$ is established from ground truth boxes.
Optimization for the first anchor shape is initialized at the highest peak in the histogram, varying $\ANCapsect$ and $\ANCscale$ to maximize the ground truth boxes within the resulting tolerance region. 
Label shapes within the optimal tolerance region are removed from the histogram, and optimization for the next anchor is initialized at the next peak.

The result is illustrated in Fig.~\ref{fig:anchors}.
Blue dots represent label shapes, the green stars represent the $10$ anchors resulting from optimization and the green boxes illustrate a \SI{30}{\percent} tolerance region.

%%%%%%%%%%%%%%%%%%%%%%%%%%%%%%%%%%%%%%%%%%%%
\mysubsection{Calculating Labels}{calc_labels}

The $3$D label arrays $\LabelIoU$, $\LabelDW$, $\LabelDL$ and $\LabelDOri$ are initialized with $0$.
By iterating through labeled objects, relevant cell locations $\left(\East, \North \right)$ occupied by label rectangles are filled.
Cells outside label boxes are $0$.
For each relevant location, the IoU of all anchors $\Anchor$ is calculated with the considered cell $\GMcell$ as the center of the rotated rectangle.
The result is stored at $\LabelIoU(\East,\North,\Anchor) := \LabelIoU(\GMcell,\Anchor)$.
The offset labels $\LabelDW$, $\LabelDL$ and $\LabelDOri$ are filled accordingly.
However, while the IoU decreases rapidly in a spatial surrounding of the true center cell, the offset labels are kept constant, since orientation and size is constant for all pixels covered by an object bounding box.

For training purposes, explained later in Section \ref{sec:loss_function}, we create a $2$D map $\LabelIoUMax \in \mathbb{R}^{\GMwidth\times\GMheight}$ containing the maximum IoU for all cells $\GMcell$ along the anchor dimension of $\LabelIoU$ by
$\LabelIoUMax(\GMcell) = \underset{\Anchor}{\max}\left(\LabelIoU(\GMcell,\Anchor)\right)$.
%

%%%%%%%%%%%%%%%%%%%%%%%%%%%%%%%%%%%%%%%%%%%%%%%%%
\mysubsection{Inferring object bounding boxes}{NetoutToBoxes}
\begin{figure}
	\vspace{2mm}
	\centering
	\includegraphics[width=0.7\linewidth]{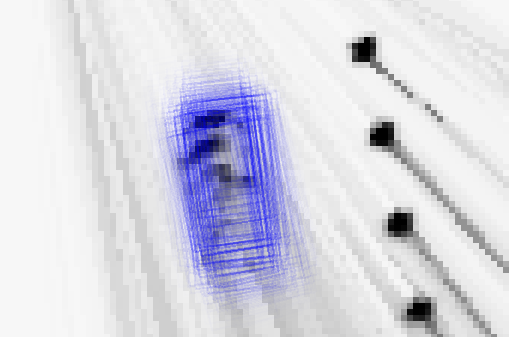}
	\caption{Network result for object bounding boxes, interpretable as a hypothesis density. Rectangles have decreasing transparency with increasing score.}
	\label{fig:hypothesis_density}
\end{figure}
For each cell $\ANCnumanchors$, boxes can be constructed using the anchors and their predicted offset.
Each resulting box comes with a score, i.e. the predicted IoU.
Fig.~\ref{fig:hypothesis_density} illustrates the result, where the bounding box transparency refers to the score.
The normalized result can be seen as a distribution of object hypotheses.

However, for many applications a single winning box is desired.
For this task, $\NetoutIoUMax(\GMcell) = \underset{\Anchor}{\max}\left(\NetoutIoU(\GMcell,\Anchor)\right)$ is calculated and boxes enclosing a higher $\NetoutIoUMax(\GMcell)$ are refused.
To speed up computation, the process starts only at local maxima in $\NetoutIoUMax(\GMcell)$, thresholded to a minimum score.
As there might be similar anchor scores for different orientations, the four best anchors in a cell are investigated first. 
Among these four, the anchor $\Anchor_{\mathrm{max}}$ with the least orientation offset $\ANCdOriRel_{\mathrm{min}}$ is considered as the winning anchor. 
Only the winning boxes were used for evaluation and in Fig.~\ref{fig:teaser}.

% *** Section - ***
\mysection{Spatial Balancing Loss Function}{loss_function}

\begin{figure}[t]
	\centering
	\vspace{1.5mm}
	\includegraphics{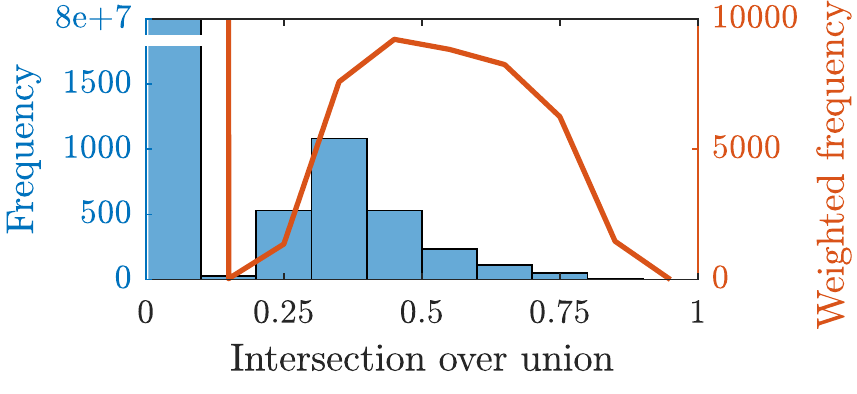}
	\caption{IoU histogram (blue) of $100$ labels $\LabelIoUMax$ and the weighted frequency of IoU occurrence (red) used for balancing during training with $\GainParam=400$ and $\FocusingParam=4$.}
	\label{fig:IoUHist}
\end{figure}

We adopt the loss function from our previous work \cite{hoermann2017GMPred} with slight modifications explained below.
Similar to \cite{hoermann2017GMPred} we face the problem of high imbalance between static background and dynamic objects where dynamic cells occur extremely rarely, as discussed in section \ref{sec:dataset_training}.
The necessity of counteracting such imbalance in the loss function when training a single stage neural network was thoroughly investigated by \cite{FocalLoss}.

We use $\LabelIoUMax$ as a spatial map to adjust weighting of the cells.
In particular $\LabelIoUMax(\GMcell) = 0$ for all background cells and $0<\LabelIoUMax(\GMcell) \leq1$ for all cells that are occupied by an object.
The weighting follows
\begin{equation}
\label{eq:loss}
L_{\Label}=\frac{\lambda_\Label}{2}\sum_\GMcell\sum_\Anchor \left( 1+\GainParam\cdot \LabelIoUMax(\GMcell)^{\FocusingParam_\Label} \right) \left(\Netout(\GMcell, \Anchor) - \Label(\GMcell, \Anchor) \right)^2
\end{equation}
where $\Label \in \left\{\LabelIoU, \LabelDW, \LabelDL, \LabelDOri  \right\} $ and
\begin{equation}
L=L_{\LabelIoU} + L_{\LabelDW}  + L_{\LabelDL}  + L_{\LabelDOri}
\end{equation}
is the total loss.
The factor $\lambda_\Label$ is used to mix the influence of output type, e.g. to weight the orientation offset similar to the anchor score (IoU).
In the term $(1+\GainParam\cdot \LabelIoUMax(\GMcell)^{\FocusingParam_\Label})$, $\GainParam$ is used to reduce unbalancing between background and object cells.
Background cells are weighted by $1$, since here $\LabelIoUMax(\GMcell)=0$, while the weight of object cells is increased up to $1+\GainParam$.
In our case, a low $\GainParam$ results in numerous false negative predictions, while choosing $\GainParam$ an order of magnitude higher than the ratio of object cells to occupied cells results in many false positives.
The parameter $\FocusingParam_\Label$ is introduced to adjust the weighing of cells within object bounds where $0<\LabelIoUMax(\GMcell)\leq1$. 
Center cells with high IoU are relatively rare compared to cells with lower IoU, as illustrated in Fig.~\ref{fig:IoUHist} where the histogram of IoU frequency among $100$ samples of $\LabelIoUMax$ is given.
Without $\FocusingParam_\Label$, i.e. $\FocusingParam_\Label=1$, the network output tends to mostly predict values at highest IoU occurrence, i.e. $\approx0.35$.
A strategy to find good training parameters is: first choose $\GainParam$ in an order of magnitude as the foreground to background ratio, and second choose $\FocusingParam_\Label$ to approximate a uniformly weighted frequency of IoU occurrence.
An example for weighted frequency is illustrated by the red curve in Fig.~\ref{fig:IoUHist}.
%

% *** Section - ***
\mysection{Dataset and Training}{dataset_training}

\begin{figure}[t]
	\centering
	\vspace{1.5mm}
	\includegraphics{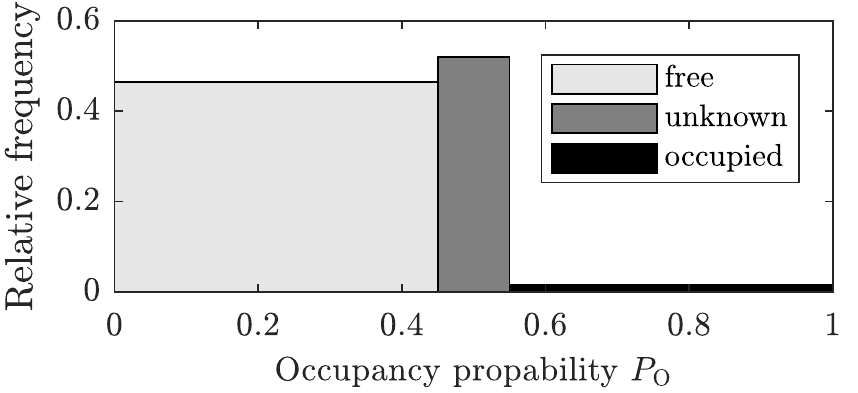}
	\caption{Occupancy histogram of input data.}
	\label{fig:POCChist}
\end{figure}

We ran the automatic label generation algorithm on about $\SI{2}{h}$ recordings of an urban shared space junction with pedestrians, bikes and motor vehicles.
The sequences were recorded at three different days observing the junction from east and west.
We used $68927$ samples for training and $1800$ samples to run evaluation experiments.

The input \DOGMA\ has spatial dimensions of $901\times901$ cells with a width of \SI{0.15}{m}. 
The histogram of $P_\Occupied$ over $100$ random samples is given in Fig.~\ref{fig:POCChist}.
The ratio of occupied to free is about $1:31$, the ratio occupied to not occupied is $1:65$.
The ratio of dynamic foreground cells to total occupied background cells is about $1:400$.
A common training strategy, hard negative mining, is to use only a sparse set of background examples for back propagation, e.g. the worst $3$ predictions, to reduce the imbalance, e.g. to $1:3$ (c.f. \cite{YOLO}).
We, however, assume about {\SI{5}{\percent}} missing labels in our dataset and therefore decided to use all cells for back propagation but employ loss balancing.
This way, the effect of a missing label vanishes in the mass of correct background labels.

The histogram in Fig.~\ref{fig:IoUHist} illustrates the extreme imbalance between $\LabelIoU=0$ and $\LabelIoU>0$, but also, a high imbalance between labels with $0<\LabelIoU<0.5$ and $\LabelIoU \geq 0.5$.
To counteract the imbalance between dynamic and static cells we chose $\GainParam=400$, according the ratio between dynamic and occupied background cells in our training data.
To reduce the imbalance within dynamic cells ($0<\LabelIoU\leq1$), $\FocusingParam_{\LabelIoU}=4$ was chosen.
The resulting weighted frequency of IoU occurrence is illustrated with a red curve in Fig.~\ref{fig:IoUHist}.
In contrast to the anchor score (IoU), yielding a local maximum at the object center cell, the offset labels have equal values in a surrounding of the center. 
Therefore, we set $\FocusingParam_{\LabelDL}=\FocusingParam_{\LabelDW}=\FocusingParam_{\LabelDOri}=1$.
The mixing parameters were chosen to $\lambda_\LabelDL=0.01$, $\lambda_\LabelDW=0.05$, $\lambda_\LabelDOri=0.25$ and $\lambda_\LabelIoU=1$.

The ADAM solver \cite{ADAM} was used for training. 
We chose the exponential decay rates to $\beta_1=0.9$ and $\beta_2=0.999$, as suggested in \cite{ADAM}. 
A base learning rate of $0.0001$ was used.
The training process was stopped after about $3.3$ epochs ($200000$ iterations) with a batch size of $1$.

% *** Results
\mysection{Results and Evaluation}{results}
\begin{figure}[t]
	\centering
	\vspace{0.8mm}
	\setlength\figureheight{3cm}
	\setlength\figurewidth{7.5cm}
\definecolor{mycolor1}{rgb}{0.00000,0.44700,0.74100}%
\begin{tikzpicture}

\begin{axis}[%
width=0.951\figurewidth,
height=\figureheight,
at={(0\figurewidth,0\figureheight)},
scale only axis,
unbounded coords=jump,
xmin=0,
xmax=1,
xlabel style={font=\color{white!15!black}},
xlabel={Recall},
ymin=0,
ymax=1,
ylabel style={font=\color{white!15!black}},
ylabel={Precision},
axis background/.style={fill=white},
xmajorgrids,
ymajorgrids,
legend style={at={(0.03,0.03)}, anchor=south west, legend cell align=left, align=left, draw=white!15!black}
]
\addplot [color=mycolor1, line width=1.5pt]
  table[row sep=crcr]{%
0.888867395299352	0.0484435510511038\\
0.883577108078924	0.0993765732876006\\
0.880499620999901	0.154507653961149\\
0.878406985975126	0.199884085896867\\
0.876905740421583	0.234639943242284\\
0.875742393576429	0.256950934579439\\
0.874721917853704	0.273419823559938\\
0.873853820598007	0.287637377658702\\
0.873014290461947	0.300515941564756\\
0.871999468049737	0.312770549864651\\
0.871249916849598	0.324059183017047\\
0.870366673321355	0.334796687614391\\
0.869521812918039	0.344509341173679\\
0.868912486268766	0.354078213806107\\
0.868386881971034	0.363765690376569\\
0.867694561923474	0.373065303609524\\
0.867210657785179	0.382287736195328\\
0.866735502781201	0.391361236859124\\
0.866366816591704	0.400425001154894\\
0.865820338531254	0.409289944556452\\
0.865093647937079	0.41835353274884\\
0.86454236384241	0.427434372064664\\
0.864025874428995	0.436282515363246\\
0.863522662842277	0.445566875473257\\
0.862853710624666	0.454542258472789\\
0.862384402230161	0.463960484957342\\
0.861547197541586	0.473387659215211\\
0.860862299465241	0.482865284391286\\
0.860220088972138	0.493097630186364\\
0.859237929418067	0.503334379413149\\
0.858588905647729	0.513192427126114\\
0.85734904172121	0.523014865473607\\
0.856705985146352	0.532769070010449\\
0.855710006394938	0.543526594834958\\
0.854827841789637	0.554455661902889\\
0.854001551381066	0.564807173287534\\
0.852879616501249	0.574835039817975\\
0.851980682854345	0.585757737583877\\
0.851231294125595	0.596948807239476\\
0.850079421406604	0.607545711456245\\
0.848946887994861	0.618939636686303\\
0.847818735204599	0.629630559811136\\
0.846903223624125	0.640889776276046\\
0.845578530576331	0.652597487397811\\
0.844254800365742	0.66354369061244\\
0.842763247052446	0.675033921302578\\
0.840908320623623	0.68523530711445\\
0.838927312177922	0.696885208966993\\
0.836448122646949	0.708181374379002\\
0.83376878965763	0.719615767586247\\
0.83096870545109	0.731898355754858\\
0.827311009984378	0.744029075804777\\
0.822889764849803	0.756608135974505\\
0.818379628055622	0.769188981913466\\
0.812506379069847	0.782452001834742\\
0.805722070844687	0.794652154926266\\
0.796801473095547	0.805869775141399\\
0.784209987370721	0.81558395456159\\
0.768739105171412	0.825994344265305\\
0.747011883968629	0.833983329509826\\
0.722607441988192	0.842073682947318\\
0.691569497471881	0.850399695470118\\
0.658390233809228	0.858684897004588\\
0.621055542114541	0.864933814681107\\
0.579458035899924	0.8718456725756\\
0.53626007224229	0.879671832269827\\
0.487845650205475	0.886077550762224\\
0.437056954289905	0.89012161297205\\
0.384297665301551	0.895513866231648\\
0.328365013677492	0.901588830043332\\
0.275164248322383	0.907742234585072\\
0.227179956365684	0.909936575052854\\
0.182439333638573	0.91261451726568\\
0.144010150495189	0.916965888689408\\
0.109904446246606	0.924377224199288\\
0.0833744797912111	0.929245283018868\\
0.0609128103837472	0.937059142702116\\
0.0419018058690745	0.943606036536934\\
0.0286762371697647	0.945348837209302\\
0.0176007900955875	0.955938697318008\\
0.0104052767098162	0.97682119205298\\
0.00581990053260908	0.970588235294118\\
0.00324503544848506	0.968421052631579\\
0.00183415047088286	0.981132075471698\\
0.00116398010652182	0.970588235294118\\
0.00049380974216077	0.933333333333333\\
0.00021163274664033	1\\
0.000176360622200275	1\\
7.05442488801101e-05	1\\
3.5272124440055e-05	1\\
3.5272124440055e-05	1\\
0	nan\\
0	nan\\
0	nan\\
0	nan\\
0	nan\\
0	nan\\
0	nan\\
0	nan\\
0	nan\\
};
\addlegendentry{precision}

\end{axis}
\end{tikzpicture}%
	\caption{Object detectio precision recall curve.}
	\label{fig:precisionrecall}
\end{figure}
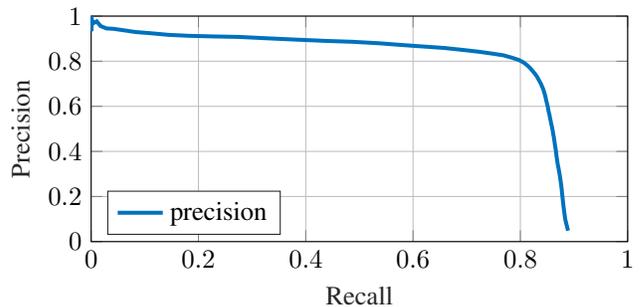
\begin{table}[t]
	\caption{Bounding box error (RMSE) and average precision (AP)}
	\label{tab:boxerrors}
	\begin{center}
		\begin{tabular}{ccccc}
			\toprule 
			position & width & length & orientation & AP\\ 
			\midrule
			%\SI{0.47}{m} & \SI{0.21}{m} & \SI{0.77}{m} & \SI{8.95}{\degree} & 0.7796\\ % V1.1 it120k
			\SI{0.47}{m} & \SI{0.21}{m} & \SI{0.76}{m} & \SI{8.72}{\degree} & 0.7594\\ % V1.1 it200k
			\bottomrule
		\end{tabular} 
	\end{center}
\end{table}
\begin{figure*}[h!]
	\vspace{1.7mm}
	\centering
	\includegraphics[width=\linewidth]{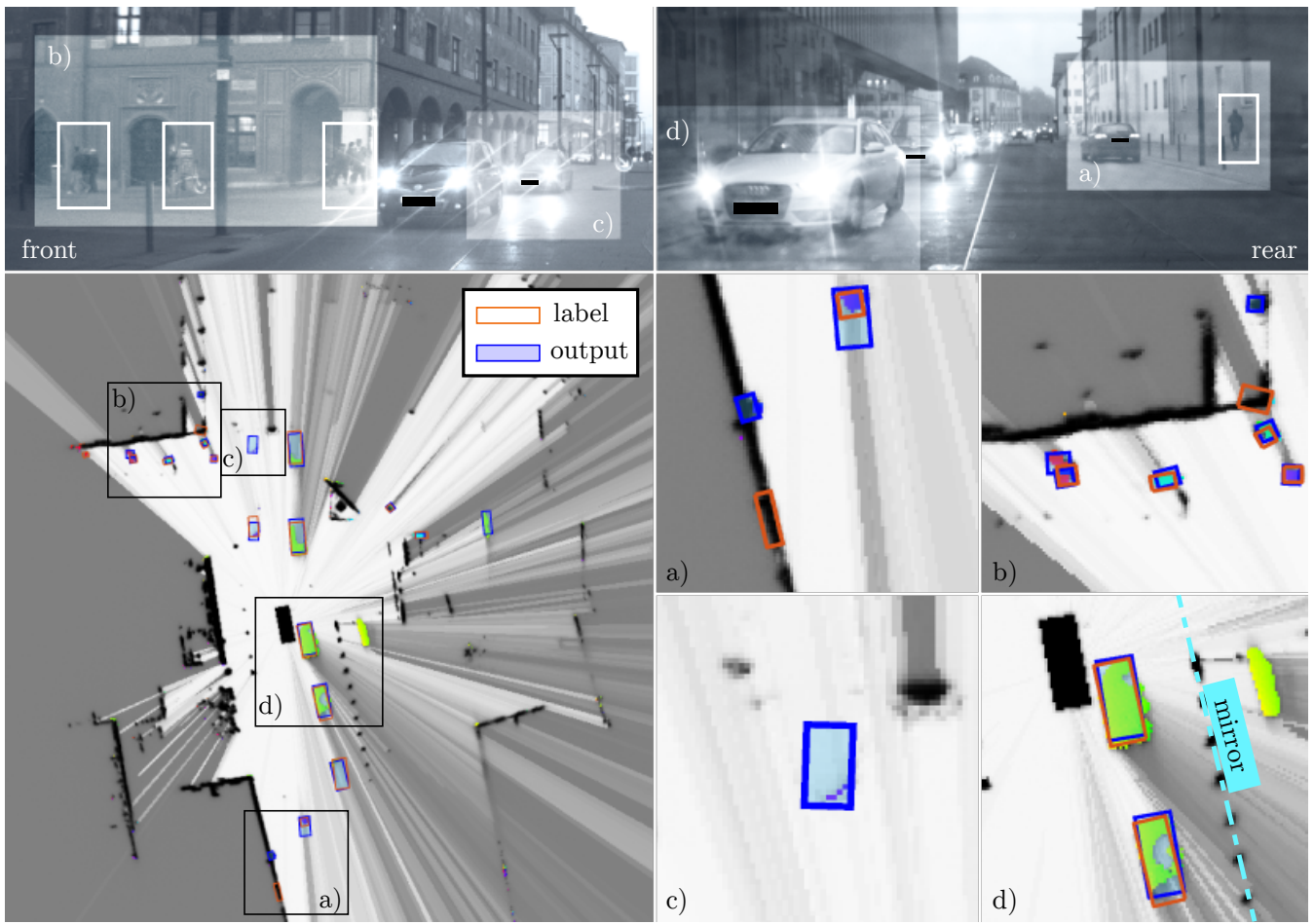}
	\caption{ 
		Example for detected and labeled objects. 
		Regions of interest (a-d) are emphasized in the camera images (top row), marked in the \DOGMA\ and shown enlarged.
		a) illustrates corrupted labels in terms of false positives, false negatives or wrong size while the network predictions are correct. 
		In b), the label algorithm fails to separate two very close pedestrians, whereas our approach yields one box per pedestrian.
		The car in c) is missed by automatic label generation but detected by the network.
		Excerpt d) illustrates a mirrored vehicle trained to be not detected using openstreetmap~\cite{OpenStreetMap}.}
	\label{fig:results}
\end{figure*}
We evaluate the object detection and pose estimation performance in a crowded downtown scenario with numerous pedestrians, bikes, cars and other road users.
The scene used for evaluation is illustrated in Fig~\ref{fig:teaser}.
It was recorded by four Velodyne VLP 16 and one $4$-layer IBEO LUX laser scanner. 
Each Velodyne provides \SI{360}{\degree} perception ranging up to \SI{100}{m} at \SI{10}{\hertz}. 
The IBEO LUX runs at \SI{12.5}{\hertz} and has a range up to \SI{200}{m} in front of the experimental vehicle with \SI{100}{\degree} opening angle.
Fusing the sensors in a \DOGMA\ covering $\SI{135.15}{m}\times\SI{135.15}{m}$ takes about \SI{30}{ms} on a GPU and is triggered at \SI{10}{\hertz}.

A video illustrating the object detection is made available online\footnote{\label{youtube}The video under \url{https://youtu.be/Rr9LOrQMgKA} illustrates the network performance.}. %%%%FOOTNOTE%%%
The network takes \SI{66.8042}{ms} on a Nvidia GTX 1080ti to process one \DOGMA\ input.
The evaluation sequences cover $1800$ example frames (\SI{3}{minutes}) including $28351$ labeled objects.
The precision recall curve for object detection performance is given in Fig.~\ref{fig:precisionrecall}.
The curve was created by varying a minimum IoU threshold $\IoUthreshold$ between $0.1$ and $1$.
A precision of 
{$0.79$} 
is achieved at recall $0.8$, while the average precision is 
{$0.7594$}. 
The prediction error, in terms of root mean square error (RMSE) over true positive object predictions with $\IoUthreshold=0.55$, of bounding box features is given in table~\ref{tab:boxerrors}.
Please note, that for the orientation error, there is a \SI{180}{\degree} ambiguity for static objects.
Therefore, the orientation error is calculated excluding 
{$328$} 
objects 
{($\approx\SI{1}{\percent}$)}
where the error was about \SI{180}{\degree}.

\refFig{results} shows the result for an example time step where predicted objects are depicted in blue and labels in orange.
It shows in particular that the network was trained not to detect objects, mirrored in high reflective building fronts.
It also shows examples where the training data contains false positives and false negatives, while the network predicts the correct result.
More examples can be seen in the video online.

% *** Conclusion ***
\mysection{Conclusions}{conclusions}

In this paper, we presented, to the best of our knowledge, the first deep learning approach to detect objects on \DOGMA s. As an object we understand a bounding box that is defined by a width, length, and orientation. A hand-engineered object tracking has been devised to bypass manual labeling of the data by using acausal information of the future movement of objects. Furthermore, we suggest a single-stage CNN that is capable of detecting the shape and orientation of objects. 

We show that our learned approach achieves similar results as the hand-engineered algorithm despite the use of solely causal information.
Furthermore, our trained network seems to have better generalization capabilities because it is able to recognize objects which the employed label algorithm lost track of and failed to reinitialize.

Since the goal of this work is to show the general potential of utilizing deep neural nets for object extraction on \DOGMA s, our dataset so far exclusively entails data recorded from a stationary platform rather than a moving one. Thus, for future work, the presented techniques should be adapted and evaluated on a moving platform.  

% *** Acknowledgements ***
\section*{Acknowledgements}

The research leading to these results has received funding from the European Union under the H2020 EU.2.1.1.7.\ ECSEL Programme, as part of the RobustSENSE project, contract number 661933.
Responsibility for the information and views in this publication lies entirely with the authors. 
The authors would like to thank all RobustSENSE partners. % for their cooperation and valuable contribution.

%%%%%%%%%%%%%%%%%%%%%%%%%%%%%%%%%%%%%%%%%%%%%%%%%%%%%%%%%%%%%%%%%%%%%%%%%%%%%%%%
\bibliographystyle{IEEEtran}
\balance
\bibliography{IEEEtranControl,SH.bib} % IEEEtranControl from where?

\end{document}